\documentclass[sigconf]{acmart}
\AtBeginDocument{
  }
\newcommand{\eg}{\textit{e}.\textit{g}.}
\newcommand{\etal}{\textit{et al}.}

\newcommand{\red}[1]{\textcolor{red}{#1}}

\definecolor{hollywoodcerise}{rgb}{0.96, 0.0, 0.63}
\definecolor{lasallegreen}{rgb}{0.03, 0.47, 0.19}
\definecolor{hanpurple}{rgb}{0.32, 0.09, 0.98}
\definecolor{green(pigment)}{rgb}{0.0, 0.65, 0.31}
\definecolor{zaffre}{rgb}{0.0, 0.08, 0.66}
\definecolor{orange}{RGB}{255,165,0}
\usepackage{pifont} 
\usepackage{graphicx}
\usepackage{multirow}

\copyrightyear{2025}
\acmYear{2025}
\setcopyright{acmlicensed}
\acmConference[MM '25] {Proceedings of the 33rd ACM International Conference on Multimedia}{October 27--31, 2025}{Dublin, Ireland.}
\acmBooktitle{Proceedings of the 33rd ACM International Conference on Multimedia (MM '25), October 27--31, 2025, Dublin, Ireland}
\acmISBN{979-8-4007-2035-2/2025/10}
\acmDOI{10.1145/3746027.3755165}
\begin{document}

\title{AuthFace: Towards Authentic Blind Face Restoration with Face-oriented Generative Diffusion Prior}

\author{Guoqiang Liang}
\authornote{Work done during internship at vivo. $^{\dagger}$Corresponding authors.}
\affiliation{
  \institution{Hong Kong University of Science and Technology (Guangzhou)}
  \city{Guangzhou}
  \country{China}
}
\email{gliang041@connect.hkust-gz.edu.cn}

\author{Qingnan Fan$^{\dagger}$}
\affiliation{
  \institution{vivo Mobile Communication Co., Ltd}
  \city{Hangzhou}
  \country{China}}
\email{fqnchina@gmail.com}

\author{Bingtao Fu}
\affiliation{
  \institution{Ant Group}
  \city{Hangzhou}
  \country{China}
}
\email{bingtaofu93@gmail.com}

\author{Jinwei Chen}
\affiliation{
 \institution{vivo Mobile Communication Co., Ltd}
 \city{Hangzhou}
 \country{China}}
\email{jinwei.chen@vivo.com}

\author{Hong Gu}
\affiliation{
  \institution{vivo Mobile Communication Co., Ltd}
 \city{Hangzhou}
 \country{China}}
\email{guhong@vivo.com}

\author{Lin Wang$^{\dagger}$}
\affiliation{
  \institution{Nanyang Technological University}
  \country{Singapore}}
\email{linwang@ntu.edu.sg}

\renewcommand{\shortauthors}{Guoqiang Liang et al.}

\begin{abstract}
Blind face restoration (BFR) is a fundamental and challenging problem in computer vision. 
To faithfully restore high-quality (HQ) photos from poor-quality ones, recent research endeavors predominantly rely on facial image priors from the powerful pretrained text-to-image (T2I) diffusion models. 
However, such priors often lead to the incorrect generation of non-facial features and insufficient facial details, thus rendering them less practical for real-world applications. 
In this paper,  we propose a novel framework, namely \textbf{AuthFace} that achieves highly authentic face restoration results by exploring a face-oriented generative diffusion prior. 
To learn such a prior, we first collect a dataset of \textit{1.5K} high-quality images, with resolutions exceeding 8K, captured by professional photographers.
Based on the dataset, we then introduce a novel face-oriented restoration-tuning pipeline that fine-tunes a pretrained T2I model. 
The photography-guided annotation system and facial prior refinements fully explore the potential of these high-quality photographic images.
In this way, the potent natural image priors from pretrained T2I diffusion models can be subtly harnessed, specifically enhancing their capability in facial detail restoration. 
Moreover, to minimize artifacts in critical facial areas, such as eyes and mouth, we propose a time-aware latent facial feature loss to learn the authentic face restoration process. 
Extensive experiments on the synthetic and real-world BFR datasets demonstrate the superiority of our approach. 
Code and models are available at \url{https://github.com/EthanLiang99/AuthFace}.
\end{abstract}




\maketitle

\section{Introduction}
Face images captured in natural settings often exhibit various forms of degradation, including compression, blur, and noise~\citep{wang2021gfpgan,wang2023dr2}. 
Capturing high-quality (HQ) face images is crucial, as humans are highly sensitive to subtle facial details. Blind face restoration (BFR) aims to reconstruct HQ images from degraded inputs and has rapidly progressed in recent years due to significant research interest. 
However, BFR remains an ill-posed problem due to the unknown degradation and the loss of valuable information resulting from these complex conditions~\citep{zhou2022codeformer}.

\begin{figure*}[h]
\centering
    \vspace{-5pt}
    \includegraphics[width=0.98\linewidth]{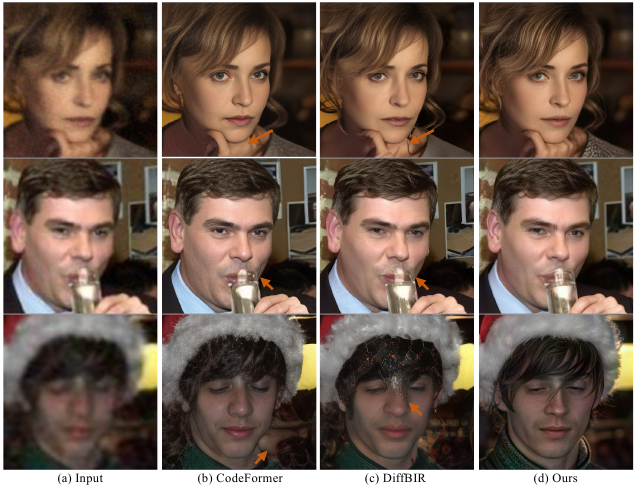}
    \vspace{-10pt}
    \caption{Compared to CodeFormer~\citep{zhou2022codeformer} and DiffBIR~\citep{lin2024diffbir}, our method more effectively captures and renders fine-grained facial details on the real-world datasets. Please zoom in for a closer view.}
    \label{fig:teaser}
\centering
\end{figure*}

\begin{figure*}[h]
\centering
    \includegraphics[width=1\linewidth]{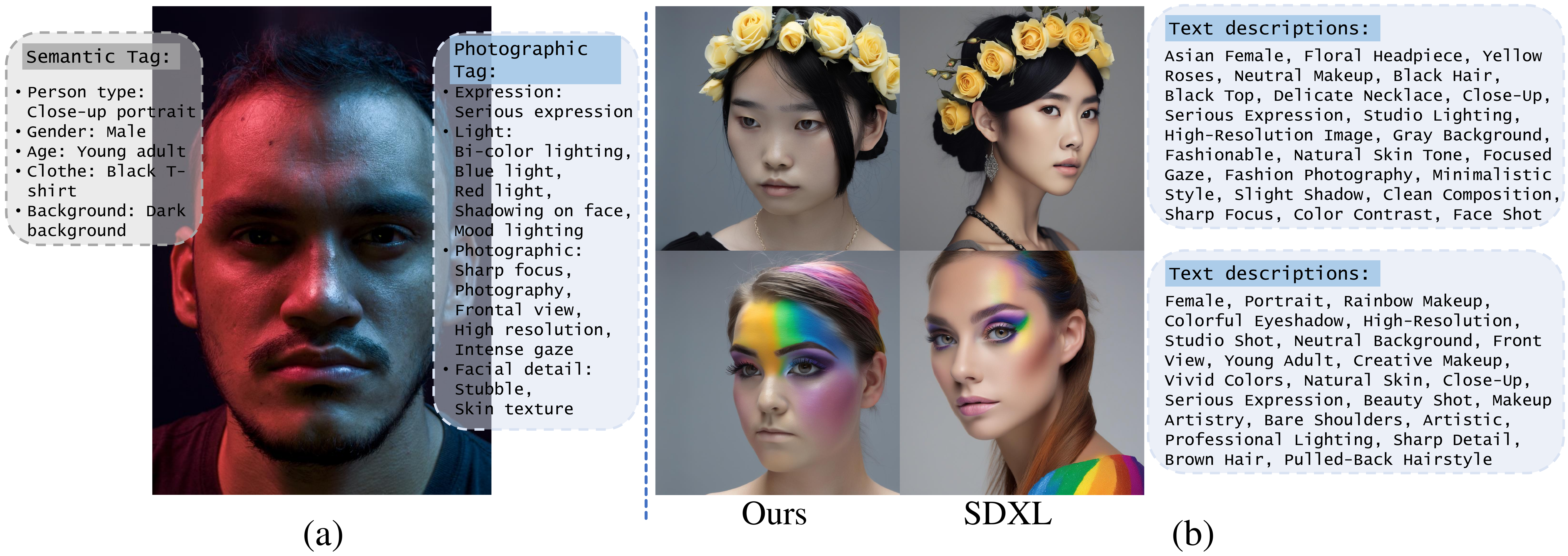}
    \vspace{-15pt}
    \caption{(a) A HQ face image with its paired tags generated through photography-guided image annotation. Specifically, we provide an additional photographic tag (\textcolor{blue}{blue box}) beyond the semantic tags used in previous methods (\textcolor{gray}{gray box}). (b) Qualitative comparison between StableDiffusion-XL (SDXL)~\citep{podell2023sdxl} and our fine-tuned model, which is exclusively trained on the collected high-quality dataset, in the T2I task. Notably, SDXL tends to generate over-smooth skin even when given prompts specifying sharp details and sharp focus. Zoom in for more details.}
    \vspace{-10pt}
    \label{fig:tag}
\centering
\end{figure*}

Sufficient prior information is critical for HQ reconstruction. 
Researchers have used geometric and reference priors from sources like~\citep{bulat2018super,kim2019progressive,ChenPSFRGAN,shen2018deep,yang2020hifacegan,yu2018face,hu2020face,zhu2022blind,ren2019face, dogan2019exemplar,li2020blind,li2020enhanced,li2018learning,chen2018fsrnet,ma2020deep} to guide face restoration. These priors, however, are limited by their sensitivity to degradation and inability to capture fine facial details, and can even result in corrupted texture details due to incorrect prior information~\citep{lu2021face}.
With advancements in generative models, such as StyleGAN~\citep{karras2020analyzing} and VQVAE~\citep{razavi2019generating}, recent works~\citep{ChenPSFRGAN,wang2021gfpgan,chan2022glean,xie2023tfrgan,wang2022panini,wang2022restoreformer,zhou2022codeformer,tsai2023dual,wang2023restoreformer++,wang2024analysis,li2020blind,li2020enhanced,li2022learning} have leveraged pretrained networks to derive facial priors, achieving superior results compared to earlier methods. Nonetheless, these approaches still face significant performance declines in handling unseen cases.
Denoising diffusion probabilistic models (DDPMs)~\citep{ho2020denoising} have shown promise as an alternative to generative adversarial networks (GANs)~\citep{song2020denoising} in image generation. 
Some approaches~\citep{yue2022difface,wang2023dr2} use pretrained DDPMs to diffuse and then denoise degraded inputs. 
However, their practical application is hindered by the loss of original identity and detailed facial features~\citep{miao2024waveface}, with pretrained DDPMs also facing limitations in representational capacity.

The remarkable success of large-scale pretrained text-to-image (T2I) models~\citep{rombach2022high,saharia2022photorealistic} has provided another promising prior.
Many researchers explore the potential of StableDiffusion (SD) models~\citep{stability_ai} as the powerful prior in challenging low-level vision tasks, including real-world image super-resolution~\citep{wang2024exploiting,lin2024diffbir,wu2023seesr,yu2024scaling} and BFR~\citep{chen2024towards,gao2024diffmac}.
Since the face details are often lost due to the degradation and down-sampling processes of VAE~\citep{rombach2022high} in SD models, BFRffusion~\citep{chen2024towards} and DiffMAC~\citep{gao2024diffmac} rely on the facial priors within SD models to recreate these details. 
However, being designed for general text-to-image tasks, SD models often fail to retain essential facial details, like skin texture (see Fig.~\ref{fig:tag} (b)).
Therefore, these methods typically produce overly smooth images in the T2I task.
Moreover, their extensive image priors can lead to the incorrect generation of non-facial features, resulting in artifacts, especially for images with ambiguous degradation.
\textit{These specific limitations -- incorrect generation of non-facial features and missing facial details (\textcolor{orange}{\textbf{orange arrows}} at Fig.~\ref{fig:teaser})-- severely limit the practical deployment of these models in real-world applications.}

To tackle these problems, we propose \textbf{Authface}, a novel BFR method with face-oriented generative diffusion prior, designed to restore highly authentic face images.
The \textbf{highlight} of our Authface is that it brings a paradigm shift for BFR -- with a two-stage training pipeline: 1) \textit{Face-oriented Fine-tuning on Pretrained T2I Model}, and 2) \textit{Highly Authentic Face Restoration.}
The underlying premise for Stage I is that pretrained T2I models~\eg, SD models, can serve as effective generative diffusion priors for restoration tasks (Sec.~\ref{Sec:finetune}). They can be customized for face-centric applications via fine-tuning while retaining their generation capabilities.

\begin{figure*}[th]
\centering
    \includegraphics[width=1\linewidth]{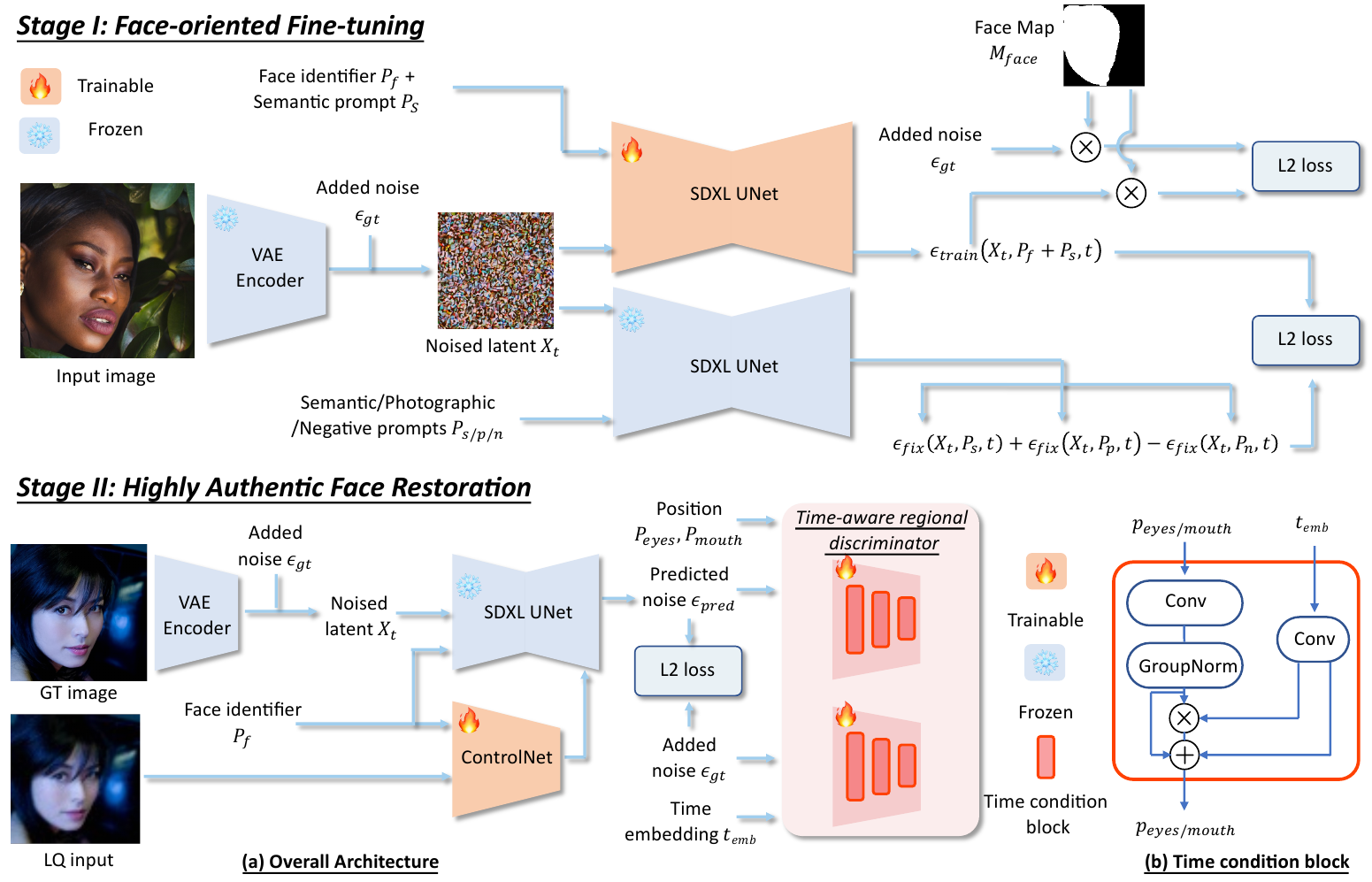}
    \caption{An overview of our framework consisting Stage I \& II. Denoising UNet, carried over from Stage I, maintains its facial priors by freezing its parameters, while ControlNet acts as an adapter for handling degraded inputs.}
    \label{fig:framework}
    \vspace{-10pt}
\centering
\end{figure*}

In analyzing key factors for fine-tuning pretrained T2I models to meet human preferences for authentic facial images, we identify three key criteria as our face-oriented generative diffusion prior: 
1)~\textbf{Quality-first image collection}. Contrary to training T2I base models with large datasets like LAION-5B~\citep{schuhmann2022laion}, the quality of the dataset, rather than its size, dictates the generation quality in the fine-tuning process.
2)~\textbf{Photography-guided image annotation}. Fine-tuning the pretrained T2I models for HQ facial tasks requires more than just basic annotations like human accessories, especially for HQ face images with a pronounced stylistic orientation (see Fig.~\ref{fig:tag} (a)). 
In line with our established criteria, we collect a curated dataset of \textbf{1.5K} HQ face images -- each enriched with detailed photographic annotations -- to fine-tune the pretrained T2I models for the first stage. With the curated dataset, we are able to fine-tune the T2I models following their original optimization strategies, as illustrated in Stage I in Fig.~\ref{fig:framework}. With fine-tuning, the pretrained T2I model is required with the detailed facial prior, which can be demonstrated with the T2I task as shown in Fig.~\ref{fig:tag} (b).
3)~\textbf{Facial Prior Refinements}. A facial prior refinement pipeline is designed to learn fine-grained features by associating a face identifier prompt with the face regions of HQ images and leveraging T2I models' inherent high-quality generative capabilities.

To achieve the goal of highly authentic face restoration in Stage II, we leverage the ControlNet~\citep{zhang2023adding} for training (Sec.~\ref{Sec:regional loss}). 
However, directly following the protocol of training ControlNet with the MSE loss tends to contribute to the loss of key facial details, such as eyes and mouths. 
To resolve this issue, we propose a \textbf{time-aware latent facial feature loss} to directly constrain the regions where humans are sensitive in the latent space. Our extensive experiments demonstrate the superior authentic detail generation performance on synthetic and real-world datasets.

\section{Related Works}

\noindent\textbf{Prior-based Blind Face Restoration}
Blind face restoration (BFR) employs a variety of priors, classified into geometric, reference, and generative categories. Geometric priors, such as facial landmarks~\citep{bulat2018super,kim2019progressive,chen2018fsrnet,ma2020deep}, face parsing maps~\citep{ChenPSFRGAN,shen2018deep,yang2020hifacegan,chen2018fsrnet}, facial component heatmaps~\citep{yu2018face}, and 3D face shapes~\citep{hu2020face,zhu2022blind,ren2019face}, provide crucial structural information for restoring degraded faces. 
Reference-based methods use images to deliver identity information, enhancing the fidelity of the restored faces~\citep{dogan2019exemplar,li2020blind,li2020enhanced,li2018learning}. Moreover, some researchers have implemented generative facial priors, like StyleGAN~\citep{karras2020analyzing}, to refine facial details~\citep{ChenPSFRGAN,wang2021gfpgan,chan2022glean,xie2023tfrgan,wang2022panini}. 
Another approach involves using pretrained Vector-Quantize codebooks that contain detailed facial information~\citep{wang2022restoreformer,zhou2022codeformer,tsai2023dual}.
Given their remarkable performance in image generation, denoising diffusion probabilistic models~\citep{ho2020denoising} have become increasingly popular in BFR. 
Notable examples, such as DifFace~\citep{yue2022difface}, DR2~\citep{wang2023dr2}, and PGDiff~\citep{yang2024pgdiff}, utilize denoising U-Nets pretrained on HQ face datasets to achieve face restoration at pixel level. 
Specifically, Zhao~\etal~\citep{zhao2023towards} attempts to improve the authentic performance via feeding network with enhanced ground-truth images.
Recently, large-scale pretrained text-to-image models like StableDiffusion (SD)\citep{stability_ai} have been employed to address the BFR problem. 
DiffBIR~\citep{lin2024diffbir} leverages SD priors for real-world image super-resolution and BFR by incorporating degraded input image information in the latent space. 
Specifically targeting BFR, BFRfusion~\citep{chen2024towards} extracts multi-scale facial features in the latent space from low-quality face images. 
\textit{However, achieving authentic BFR with pretrained T2I models in the latent space remains underexplored.}

\noindent\textbf{Fine-Tuning}
Fine-tuning is widely used to align pretrained large language models (LLMs) with human preferences, improving their effectiveness~\citep{betker2023improving}. 
This technique, successful in LLMs with small, HQ datasets~\citep{touvron2023llama,zhou2024lima}, has been adapted to text-to-image models to enhance text-image alignment~\citep{dai2023emu,li2024playground,li2024cosmicman}. 
For example, Emu~\citep{dai2023emu} improves aesthetic alignment using fine-tuned HQ image-text pairs. 
Playground v2.5~\citep{li2024playground} enhances human features using a quality-controlled dataset, and CosmicMan~\citep{li2024cosmicman} generates superior human-centric content with large, refined datasets.
Recently, RAP-SR~\cite{wang2024rap} demonstrated the potential of restoration-oriented prompt optimization by fine-tuning pretrained text-to-image models with carefully curated positive and negative samples.
\textit{However, these methods often produce overly smooth images, which may not be ideal for BFR tasks where authentic and realistic images are essential.}

\vspace{-5pt}
\section{Methodology}
\vspace{-2pt}
The goal of our work is to achieve authentic face restoration by minimizing unrealistic outcomes and enhancing the rendition of human-preferred features. It is structured into two distinct stages:
\textbf{1)} \textit{Face-oriented Tuning on Pre-trained T2I Model}. We integrate supervised fine-tuning~\citep{ouyang2022training} and quality-tuning~\citep{dai2023emu} strategies to refine StableDiffusion-XL (SDXL), enhancing it with detailed facial features as our face-oriented generative diffusion prior (Sec.~\ref{Sec:finetune});
\textbf{2)}\textit{ Highly Authentic Face Restoration}. Utilizing the face-oriented generative diffusion prior, we implement ControlNet~\citep{zhang2023adding} to direct the restoration process based on the quality of input degradation (Sec.\ref{Sec:regional loss}). 
Moreover, we introduce a time-aware latent facial feature loss to improve key facial features during restoration (Sec.\ref{Sec:regional loss}).

\vspace{-9pt}
\subsection{Stage I: Face-oriented Fine-tuning on Pre-trained T2I Model}
\label{Sec:finetune}
In the following section, we introduce the two main components of our face-oriented tuning procedure: (1) photography-guided data annotation that goes beyond basic semantic labels, and (2) facial prior refinements that leverage our high-quality dataset to enhance the original SDXL model. \textit{Please refer to the supplementary material for details on how we collected the HQ dataset under a quality-first principle.}

\noindent \textbf{Photography-guided Image Annotation.}
The quality of prompts is essential for both training~\citep{betker2023improving} and fine-tuning~\citep{li2024cosmicman} pretrained T2I models. 
For example, CosmicMan~\citep{li2024cosmicman} fine-tunes SDXL for human-centric content generation by breaking human parsing maps into several parts to provide detailed annotations. 
However, for face-oriented tuning tasks, densely annotated images are less effective. 
After cropping and alignment, the semantic information in facial images is limited, often capturing only overall human attributes. 
This differs significantly from other tasks where densely packed semantic information is prevalent. 
For face-oriented tuning, capturing stylistic information beyond basic semantics is crucial. 
In portrait photography, this includes expressions, skin texture, makeup, and lighting, essential for authentic face restoration. 

Therefore, we apply photography-guided data annotation to generate prompts for our fine-tuning dataset, especially given that our dataset consists of HQ portraits by professional photographers with strong stylistic tendencies.
We follow previous methods~\citep{betker2023improving,li2024cosmicman} to realize automatic captioning tasks with Vision-Language Models (VLMs).
Specifically, we leverage the pretrained 
Qwen2.5-VL-7B-Instruct~\citep{bai2025qwen2} as the automatic caption to generate a tag-style prompt to avoid redundant prepositions and adverbs~\citep{hertz2022prompt}.
Fig.~\ref{fig:tag} (a) illustrates some examples of photography-guided data annotation. 

\noindent \textbf{Facial Prior Refinements.}
Our primary goal in refining the facial prior is to enhance the generation of authentic facial details while suppressing common artifacts in Blind Face Restoration (BFR). To achieve this, we propose a facial prior refinement pipeline designed to learn fine-grained features (e.g., skin texture, hair strands) from high-quality datasets. Concurrently, this pipeline leverages the robust generative capabilities of the pre-trained SDXL model to mitigate the generation of low-quality or unrealistic outputs.
While techniques like DreamBooth~\cite{ruiz2023dreambooth} have gained popularity for personalizing generative models to specific subjects or styles within the generative AI community, their direct application often falls short for BFR tasks. Specifically, there remains a gap in methods that explicitly link textual prompts to the synthesis of fine-grained facial attributes, such as realistic skin textures and hair details, which are crucial for high-fidelity restoration.

To address this, we first associate a dedicated positive prompt, denoted $\mathbf{P}_f$ (e.g., "A hyper-realistic portrait"), with high-quality face images during training. This prompt guides the model towards generating outputs with desired aesthetic qualities. Crucially, to focus the learning process on intricate facial details, we employ a spatially masked loss function. We generate a binary face mask $\mathbf{M}_{face}$ from corresponding face parsing maps for each training image.
The L2 loss between added noise $\epsilon_{gt}$ and predicted noise from trainable UNet $\epsilon_{train}(\mathbf{X}_t, \mathbf{P}_{f} + \mathbf{P}_{s}, \mathbf{t})$ is then calculated only within the facial regions defined by $\mathbf{M}_{face}$:
\begin{equation}
\mathcal{L}_{mask} = \| (\epsilon_{gt} - \epsilon_{train}(\mathbf{X}_t, \mathbf{P}_f + \mathbf{P}_{s}, t)) \odot \mathbf{M}_{face} \|^2 ,
\label{eq:masked_loss}
\end{equation}
where $\mathbf{X}_t$ is the noised input at timestep $t$, $\mathbf{P}_{s}$ is the sematic prompt, and $\odot$ denotes element-wise multiplication.

Furthermore, inspired by Classifier-Free Guidance (CFG)~\cite{ho2022classifier}, which is typically employed during inference to enhance prompt adherence and sample quality, we introduce a training objective that incorporates negative guidance. This mechanism aims to actively steer the model during training away from generating undesirable features commonly observed in prior BFR results (e.g., blurriness, unrealistic textures, artifacts). We define a set of negative prompts, $\mathbf{P}_n$, describing these unwanted characteristics. We then formulate a guidance loss term: 
\begin{equation}
\begin{aligned}
    \mathcal{L}_{cfg} =  \| \epsilon_{fix}(\mathbf{X}_t, \mathbf{P}_s, \mathbf{t}) + \epsilon_{fix}(\mathbf{X}_t, \mathbf{P}_p, \mathbf{t}) - \epsilon_{fix}(\mathbf{X}_t, \mathbf{P}_n, \mathbf{t}) \\  - \epsilon_{train}(\mathbf{X}_t, \mathbf{P}_f + \mathbf{P}_{s}, t)) \|^2,
\label{eq:cfg}
\end{aligned}
\end{equation}
where $\epsilon_{fix}$ is predicted noise from the frozen UNet.

The total loss for this Face-oriented Fine-tuning stage (Stage I) is a weighted sum of the masked reconstruction loss and the negative guidance term:

\begin{equation}
\mathcal{L}_{stage1} = \mathcal{L}_{mask} + \lambda_{cfg} \mathcal{L}_{cfg},
\label{eq:stage1_total_loss}
\end{equation}
where $\lambda_{cfg}$ is a hyperparameter balancing the two objectives.

\begin{figure*}[h]
\centering
    \includegraphics[width=0.95\linewidth]{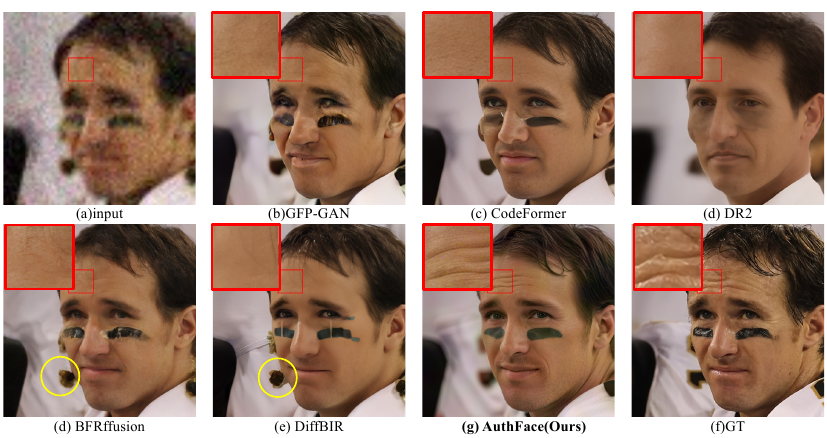}
    \vspace{-5pt}
    \caption{Qualitative results on CelebA-Test dataset. \red{Red box areas} highlight the detailed skin texture achieved by our method.}
    \vspace{-5pt}
    \label{fig:celeba}
\centering
\end{figure*}

\subsection{Stage II: Highly Authentic Face Restoration}

Fig.~\ref{fig:framework} illustrates the structure of stage II. Given the fine-tuned SDXL model as our face-oriented generative diffusion prior, we need an adaptor that can control the fine-tuned SDXL to generate high-quality facial images based on its degraded input. With the successful application of ControlNet~\citep{zhang2023adding} in real-world image super-resolution~\citep{wu2023seesr,yu2024scaling}, we apply it as the controller for the BFR task.

The training of stage II is as follows. The latent representation of an HQ facial image is obtained by the encoder of a pretrained VAE, denoted as $\mathbf{z}_0$. The diffusion process progressively introduces noise to $\mathbf{z}_0$, resulting in a noisy latent $\mathbf{z}_t$, where $t$ represents the randomly sampled diffusion step. The restoration is conditioned on the additional input $\mathbf{c}$, which is the degraded face image, guiding the generation process. For each diffusion step $t$, the noisy latent $\mathbf{z}_t$ is processed together with the control condition $\mathbf{c}$, and Face identifier $\mathbf{P}_f$  ["A hyper-realistic portrait"]. 
We train the ControlNet by minimizing the $L_2$ loss between the predicted noise $\epsilon_{pred}$ and the added noise $\epsilon_{gt}$ ($\epsilon \sim \mathcal{N}(0, I)$).
The optimization objective is:
\begin{equation}
    \mathcal{L}_{noise} = \| \epsilon_{gt} - \epsilon_{pred}(\mathbf{z}_t, t, {P}_f, \mathbf{c}) \|^2 .
\label{eq:mseloss}
\end{equation}

Specifically, we freeze the parameters of our fine-tuned SDXL model to preserve the enhanced facial priors and its original natural image priors. We initialize  ControlNet with the encoder from our fine-tuned SDXL model while solely training ControlNet.

\noindent \textbf{Time-aware Latent Facial Feature Loss.}
\label{Sec:regional loss}
Reducing incorrect generation is crucial for authentic face restoration, as viewers are particularly sensitive to key facial features such as the eyes and mouth. However, the MSE loss (Eq.\ref{eq:mseloss}) used to train ControlNet imposes only a holistic constraint, with both the background and face in the degraded image contributing equally to the optimization process. Thanks to the spatially invariant properties of ControlNet’s conditioning embedding module, the latent space preserves spatial dimensions\citep{avrahami2023blended}. Consequently, we can precisely locate the facial and mouth regions in the ground-truth (GT) image using a pretrained face landmark detection network~\citep{zheng2021farl} and then map these pixel-level coordinates to their corresponding latent space locations ($P_{eyes}$, $P_{mouth}$) by downsampling with a factor of eight. With these latent space positions for the eyes and mouth, we crop the corresponding regions from both the predicted noise $\epsilon_{pred}$ and the added noise $\epsilon_{gt}$, thereby obtaining the facial feature patches $P_{eye} = \{p_{eye}, p_{eye}^{pred}\}$ and $P_{mouth} = \{p_{mouth}, p_{mouth}^{pred}\}$.

To enhance key facial features, we propose a time-aware latent facial feature loss that provides additional constraints on the eyes and mouth using separate time-aware regional discriminators. Unlike GFP-GAN~\citep{wang2021gfpgan}, our method integrates the diffusion and denoising processes of DDPMs by explicitly treating time as a variable. Previous studies~\citep{wang2024exploiting, avrahami2023blended, choi2022perception} have demonstrated that the noise distribution varies significantly during denoising, with generated results evolving from coarse shapes to high-resolution details. Ignoring these distribution differences under various noise disturbances poses a significant challenge for optimizing discriminators, as also noted by \cite{he2024one}.
To incorporate time embedding guidance, our time-aware regional discriminator is designed with dedicated time condition blocks, as depicted in Fig.\ref{fig:framework} Stage II (b). Specifically, we utilize the time embedding provided by the SDXL UNet and employ spatial feature transfer\citep{wang2018recovering} along with group normalization~\citep{wu2018group} to endow the facial features $p_{eye/mouth}$ with time information.

The time-aware latent facial feature loss is defined as follows:
\begin{equation}
\begin{aligned}
\mathcal{L}_{facial} = \sum_{P \in {P_{\text{eye}}, P_{\text{mouth}}}}  \Big( & \lambda_d \mathbb{E}_{p^{\text{pred}}} \left[ \log(1 - \mathbb{D}_P(p^{\text{pred}})) \right] \\
& + \lambda_s \left| \text{Gram}(\psi(p^{\text{pred}})) - \text{Gram}(\psi(p)) \right| \Big),
\end{aligned}
\end{equation}
where $\mathbb{D}_P$ refers to the discriminators for different facial regions, specifically $\mathbb{D}_{eyes}$ and $\mathbb{D}_{mouth}$. $\psi$ represents multi-scale features of the regional facial feature discriminator. Gram() operation refers to calculating the Gram matrix static~\citep{gatys2016image}
$\lambda_d$ and $\lambda_s$ are the weights of the discriminative loss and the style loss, respectively.

The total loss function of this Highly Authentic Face Restoration (Stage II) is defined as $\mathcal{L}_{\text{total}} = \mathcal{L}_{noise}  + \mathcal{L}_{facial}$.

\section{Experiments}
\subsection{Experimental Settings}
\noindent \textbf{Implementation Details:}
Our base model is initialized from SDXL, and we fine-tune the entire U-Net from this base model.
We employ AdamW~\citep{loshchilov2017decoupled} optimizer with the learning rate of $5e-7$ during the finetuning process, where the batch size and the training iteration are set to 96 and 20k, respectively. 
We apply the same optimizer with the learning rate of $2e-5$ with the batch size 48 for training ControlNet.
All experiments are conducted on four NVIDIA A100 GPUs in the resolution of 1024 $\times$ 1024 for finetuning  model and 512 $\times$ 512 for training ControlNet.

\noindent \textbf{Training and Test Dataset:}
The training dataset for the fine-tuning process of our face-oriented model comprises \textbf{1.5K} high-quality face images, each enriched with detailed photographic annotations. 
For training our AuthFace network, we resize the FFHQ dataset~\citep{karras2019style} from a resolution of 1024×1024 to 512×512. 
To form training pairs, we follow the settings, including degradation types and degrees, as outlined in previous methods~\citep{wang2021gfpgan, chen2024towards}. 
Following~\citep{zhou2022codeformer, zhao2023towards, yang2024pgdiff}, we evaluate our method on a synthetic dataset, CelebA-Test~\citep{liu2015deep}, and three real-world datasets: LFW-Test~\citep{wang2021gfpgan}, WebPhoto-Test~\citep{wang2021gfpgan}, and WIDER-Test~\citep{zhou2022codeformer}.

\noindent \textbf{Metrics:}
To evaluate our method's performance on the Celeb-A dataset with ground truth, we use PSNR~\citep{hore2010image}, SSIM~\citep{wang2004image}, and LPIPS~\citep{zhang2018unreasonable}. We incorporate non-reference image quality metrics including MUSIQ~\citep{ke2021musiq} and ManIQA~\citep{yang2022maniqa}. Specifically, we employ TOPIQ~\citep{chen2024topiq}, \textbf{which is trained on a face image quality assessment (IQA) dataset}, as our face IQA metric.

\subsection{Comparison and Evaluation}
We compare our method with SOTA BFR methods in three different categories: \textbf{(I)} GAN-based methods, including GFP-GAN~\citep{wang2021gfpgan} and PSFR-GAN~\citep{ChenPSFRGAN}; 
\textbf{(II)} Codebook-based method, including CodeFormer~\citep{zhou2022codeformer};
\textbf{(III)} Diffusion-based methods, including DR2~\citep{wang2023dr2} and BFRffusion~\citep{chen2024towards};
Notably, we compare our method with SOTA IR methods, StableSR~\cite{wang2024exploiting} and DiffBIR~\citep{lin2024diffbir} with their face versions.
All methods are tested with official codes.

\begin{figure*}[t]
\centering
    \includegraphics[width=1\linewidth]{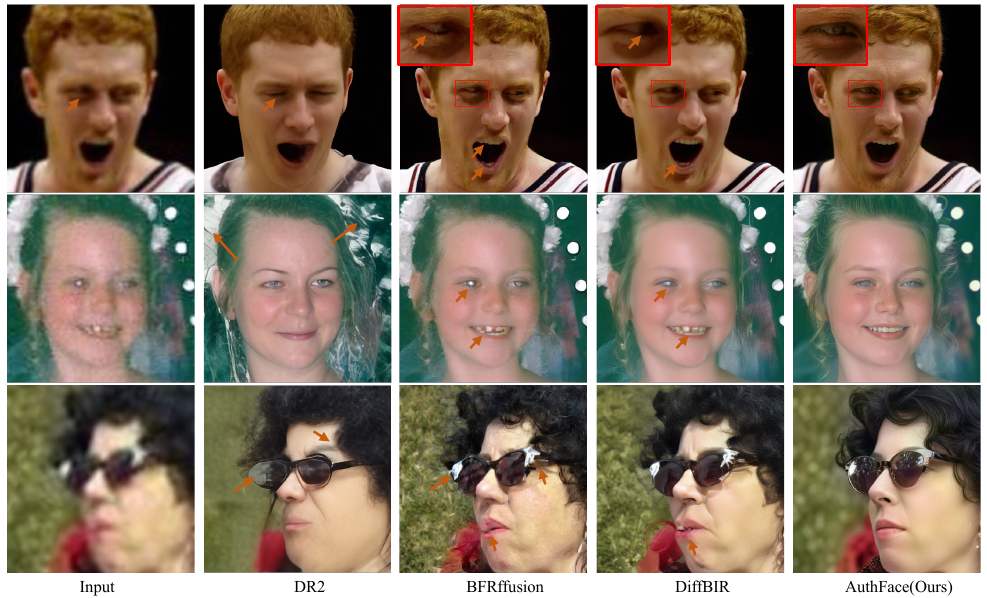}
    \vspace{-15pt}
    \caption{Qualitative results are presented on LFW-Test~\citep{wang2021gfpgan}, WebPhoto-Test~\citep{wang2021gfpgan}, and WIDER-Test~\citep{zhou2022codeformer}. Zoom in for details.}
    \vspace{-5pt}
    \label{fig:real-world}
\centering
\end{figure*}

\begin{table*}[t]
\centering
\setlength{\tabcolsep}{0.012\linewidth}
\begin{tabular}{c|c|ccc|ccccc}
\hline
Datasets & \multicolumn{1}{c|}{Metrics} & \multicolumn{1}{c}{\begin{tabular}[c]{@{}c@{}}GFPGAN \end{tabular}} & \multicolumn{1}{c}{\begin{tabular}[c]{@{}c@{}}PSFRGAN \end{tabular}} & \multicolumn{1}{c|}{\begin{tabular}[c]{@{}c@{}}CodeFormer \end{tabular}} & \multicolumn{1}{c}{\begin{tabular}[c]{@{}c@{}}DR2 \end{tabular}} & \multicolumn{1}{c}{\begin{tabular}[c]{@{}c@{}}BFRffusion \end{tabular}} &  \multicolumn{1}{c}{\begin{tabular}[c]{@{}c@{}}DiffBIR \end{tabular}}  &  \multicolumn{1}{c}{\begin{tabular}[c]{@{}c@{}}StableSR \end{tabular}} & Ours \\ \hline
\multirow{7}{*}{CelebA} & PSNR$\uparrow$ & 24.65 & \textcolor{blue}{24.68} & \textcolor{red}{\textbf{25.15}} & 21.43 & \textcolor{red}{\textbf{26.19}} & \textcolor{blue}{25.58} & 25.32 & 25.22 \\ 
 & SSIM$\uparrow$ & \textcolor{red}{\textbf{0.6669}} & 0.6322 & \textcolor{blue}{0.6647} & 0.5943 & \textcolor{red}{\textbf{0.6829}} & 0.6737 & \textcolor{blue}{0.6762} & 0.6364 \\ 
 & LPIPS$\downarrow$ & \textcolor{blue}{0.2308} & 0.2943 & \textcolor{red}{\textbf{0.2269}} & 0.3443 & 0.2272 & \textcolor{blue}{0.1969} & \textcolor{red}{\textbf{0.1833}} &0.2104 \\ 
 & MANIQA$\uparrow$ & \textcolor{red}{\textbf{0.5388}} & 0.4838 & \textcolor{blue}{0.5374} & 0.5151 & 0.5571 & \textcolor{blue}{0.6179} & 0.5654 & \textcolor{red}{\textbf{0.6246}} \\ 
 & MUSIQ$\uparrow$ & \textcolor{blue}{73.91} & 73.39 & \textcolor{red}{\textbf{75.56}} & 70.37 & 71.70 & \textcolor{blue}{74.89} & \textcolor{red}{\textbf{75.46}} & 74.66 \\ 
 & NIQE$\downarrow$ & \textcolor{blue}{4.080} & \textcolor{red}{\textbf{3.956}} & 4.590 & 4.921 & 4.901 & 5.365  & \textcolor{blue}{4.658} & \textcolor{red}{\textbf{4.509}} \\ 
 & TOPIQ-FACE$\uparrow$ & \textcolor{blue}{0.7586} & 0.7320 & \textcolor{red}{\textbf{0.7976}} & 0.7322 & 0.7286 & \textcolor{blue}{0.8194} & 0.8136 & \textcolor{red}{\textbf{0.8463}} \\ \hline
\multirow{4}{*}{LFW} & MANIQA$\uparrow$ & \textcolor{red}{\textbf{0.5293}} & 0.4881 & \textcolor{blue}{0.5260} & 0.5075 & 0.5169 & \textcolor{blue}{0.5955} & 0.5461 & \textcolor{red}{\textbf{0.6036}} \\ 
 & MUSIQ$\uparrow$ & \textcolor{blue}{73.47} & 73.43 & \textcolor{red}{\textbf{75.39}} & 70.92 & 69.59 & \textcolor{blue}{74.50} & \textcolor{red}{\textbf{75.43}} & 73.83 \\ 
 & NIQE$\downarrow$ & \textcolor{red}{\textbf{3.902}} & \textcolor{blue}{3.976} & 4.448 & 4.989 & 4.912 & 5.677 & \textcolor{blue}{4.461} & \textcolor{red}{\textbf{4.239}}  \\ 
 & TOPIQ-FACE$\uparrow$ & 0.7454 & \textcolor{blue}{0.7525} & \textcolor{red}{\textbf{0.7825}} & 0.7328 & 0.6824 & 0.7839 & \textcolor{blue}{0.7875} & \textcolor{red}{\textbf{0.8265}} \\ \hline
\multirow{4}{*}{WebPhoto} & MANIQA$\uparrow$ & \textcolor{red}{\textbf{0.5114}} & 0.4450 & \textcolor{blue}{0.5033} & 0.4585 & 0.4397 & \textcolor{blue}{0.5654} & 0.5198 & \textcolor{red}{\textbf{0.5742}}\\ 
 & MUSIQ$\uparrow$ & \textcolor{blue}{72.11} & 71.67 & \textcolor{red}{\textbf{73.99}} & 67.17 & 62.08 & 72.77 & \textcolor{blue}{73.92} & \textcolor{red}{\textbf{74.34}}\\ 
 & NIQE$\downarrow$ & \textcolor{blue}{4.149} & \textcolor{red}{\textbf{3.983}} & 4.626 & 4.993 & 5.556 & 6.007 & \textcolor{blue}{4.725} & \textcolor{red}{\textbf{4.325}} \\ 
 & TOPIQ-FACE$\uparrow$ & \textcolor{blue}{0.6990} & 0.6695 & \textcolor{red}{\textbf{0.7244}} & 0.6463 & 0.5630 & 0.7276 & \textcolor{blue}{0.7360} & \textcolor{red} {\textbf{0.7767}} \\ \hline
\multirow{4}{*}{WIDER} & MANIQA$\uparrow$ & \textcolor{red}{\textbf{0.5055}} & 0.4614 & \textcolor{blue}{0.4959} & 0.4746 & 0.4704 & \textcolor{blue}{0.5924} & 0.5223 & \textcolor{red}{\textbf{0.6072}} \\ 
 & MUSIQ$\uparrow$ & \textcolor{blue}{72.81} & 71.52 & \textcolor{red}{\textbf{73.42}} & 67.21 & 61.87 & 73.24 & \textcolor{blue}{73.58} & \textcolor{red}{\textbf{74.56}} \\ 
 & NIQE$\downarrow$ & \textcolor{red}{\textbf{3.809}} & \textcolor{blue}{3.889} & 4.119 & 5.008 & 4.774 & 5.588 & \textcolor{blue}{4.156} & \textcolor{red}{\textbf{3.971}}  \\ 
 & TOPIQ-FACE$\uparrow$ & \textcolor{blue}{0.7213} & 0.7096 & \textcolor{red}{\textbf{0.7444}} & 0.6962 & 0.5894 & \textcolor{blue}{0.7768} & 0.7588 & \textcolor{red}{\textbf{0.8411}} \\ \hline
\end{tabular}
\vspace{5pt}
\caption{Quantitative results for blind face restoration on both synthetic and real-world datasets. The best result is highlighted in \textcolor{red}{\textbf{red}} while the second-best result is highlighted in \textcolor{blue}{blue}.}
\vspace{-15pt}
\label{tab:result}
\end{table*}
\begin{figure*}[h]
    \includegraphics[width=0.95\textwidth]{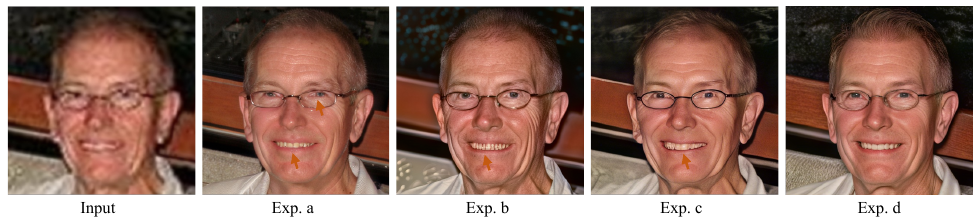}
    \vspace{-5pt}
    \caption{Visualization of ablation results. Zoom in for more details.}
    \vspace{-5pt}
    \label{fig:ablation}
\end{figure*}
\noindent \textbf{Comparison on Synthetic Dataset:}
Quantitative results in Table~\ref{tab:result} show our method's superior performance on the CelebA-Test dataset. 
It surpasses all other diffusion-based methods on non-reference IQA metrics, with the exception of MUSIQ.
Notably, we achieve state-of-the-art performance in TOPIQ-FACE, a metric that evaluates image quality by detecting faces and masking non-face regions, providing strong validation for our approach to the BFR task.
Qualitatively, as illustrated in Fig.~\ref{fig:celeba}, our AuthFace model delivers authentic face restoration. 
All methods except ours fail to recover the face paint, and our approach yields the best hair details. The red-boxed areas offer zoomed-in views of the skin texture, where our method not only reveals fine skin pores but also restores the wrinkles on the forehead.
Moreover, even though DiffBIR employs negative prompts such as "blurry, low-quality," it struggles to reconstruct clear facial images in this case.
In contrast, only our method produces realistic results in these critical regions, delivering superior details in the hair and skin texture.

\noindent \textbf{Comparison on Real-world Dataset:}
The robustness of our method is validated by its state-of-the-art performance across all metrics and real-world datasets—except for the MUSIQ score in the LFW-Test and WebPhoto-Test datasets, as shown in Table~\ref{tab:result}. Notably, our TOPIQ-FACE score in the WIDER-Test dataset exceeds the baselines by \textbf{0.0643}.
On the LFW-Test dataset, BFRfusion and DiffBIR fail to reconstruct realistic eye regions and exhibit noticeable artifacts (see the \textcolor{red}{red} box in the first row of Fig.\ref{fig:real-world}), while DR2 mistakenly restores closed eyes. 
In contrast, as shown in the second row of Fig.\ref{fig:real-world}, our method accurately reconstructs both teeth and eyes without introducing artifacts around the hair, as highlighted by the orange arrows.
On the WIDER-Test dataset, our approach not only precisely reconstructs fine details such as sunglasses but also effectively avoids artifacts around the mouth (see the \textcolor{red}{red} box areas). 
\textit{More visualization results are provided in the appendix and supplementary materials.}

\subsection{Ablation Study}
We conduct experiments to demonstrate the effectiveness of the proposed face-oriented fine-tuning and time-aware latent facial feature loss, as shown in Table~\ref{tab:ablation} and Fig.~\ref{fig:ablation}. \textit{In addition, we present qualitative and quantitative ablation study results for the losses used in face-oriented fine-tuning on the text-to-image task; please refer to the supplementary material for further details.}

\noindent \textbf{Effectiveness of Face-oriented Fine-Tuning:}
We conducted an ablation study to evaluate the effectiveness of face-oriented tuning on WIDER-Test, as shown in Tab.~\ref{tab:ablation} (a) and (b). In experiment (a), the original SDXL is used as the base model, and ControlNet is initialized with it. In experiment (b), the fine-tuned SDXL is used as the base model, and ControlNet is initialized with this fine-tuned version.
Except for the MANIQA score, experiment (b) consistently outperforms experiment (a), highlighting the necessity of face-oriented tuning for the generative diffusion prior. 
Experiment (b) enhances facial details such as eyebrows and skin texture (red box in Fig.~\ref{fig:ablation}) and eyelashes (red box), which align the great improvemnt on the TOPIQ-FACE score.
Additionally, it reduces errors in key facial features, resulting in clearer eyes than experiment (a)'s. However, it also results in the artifacts in the teeth areas.

\noindent \textbf{Effectiveness of Time-aware Latent Facial Feature Loss:}
To evaluate the effectiveness of the time-aware latent facial feature loss, we conducted experiments as shown in Table~\ref{tab:ablation} (b), (c), and (d). Using a discriminator without time embedding (experiment (c)) negatively impacted optimization, leading to performance drops across most metrics compared to experiment (b), with the exception of the MANIQA score. This result demonstrates the necessity of incorporating additional time embedding information to better optimize the regional discriminators. By modulating regional noise features with time embedding, our time-aware loss achieves the best performance on this real-world dataset.

As shown in Fig.~\ref{fig:ablation}, incorporating the latent space facial feature loss in experiment (d) significantly improves the restoration of the eyes and mouth compared to experiment (a) (see the orange arrows). Notably, although experiment (c) produces some improvement in the restoration quality of the teeth regions, it also compromises the detailed skin texture observed in experiment (b). In contrast, the time-aware strategy in experiment (d) not only reduces artifacts (as indicated by the orange arrows) but also preserves fine details, such as delicate skin texture and sharp hair.

\begin{table}[t]
\small
\centering
\setlength{\tabcolsep}{0.01\linewidth}
\begin{tabular}{c|cc|cc|cccc}
\hline
\multirow{2}{*}{Exp.} & \multicolumn{2}{c|}{Diffusion Prior} & \multicolumn{2}{c|}{$\mathcal{L}_{facial}$} & \multicolumn{3}{c}{Metrics}      \\
                         &SDXL           & Ours           & Const.         & Time-aware           & MANIQA$\uparrow$ & MUSIQ$\uparrow$ & TOPIQ$\uparrow$ \\ \hline
(a) & \ding{51}      &                &                  &                 & 0.5381 & 72.25 & 0.6962  \\
(b)            &    & \ding{51}      &                  &                 & 0.5597 & \textcolor{blue}{74.02} & \textcolor{blue}{0.7868}  \\
(c)             &   & \ding{51}      & \ding{51}        &                 & \textcolor{blue}{0.5802} & 73.66 & 0.7721  \\
(d)             &   & \ding{51}      &                  & \ding{51}       & \textcolor{red}{\textbf{0.6072}} & \textcolor{red}{\textbf{74.56}} & \textcolor{red}{\textbf{0.8411}}  \\ \hline
\end{tabular}
\caption{Ablation studies of variant generative diffusion prior and time-aware latent facial feature loss. The highest result is highlighted in \textcolor{red}{\textbf{red}} while the second highest result is highlighted in \textcolor{blue}{blue}.}
\vspace{-20pt}
\label{tab:ablation}
\end{table}

\section{Conclusion}
This paper presented a new approach for achieving authentic face restoration by avoiding incorrect generations and enhancing facial details. 
Specifically, we proposed a face-oriented restoration-tuning paradigm to fine-tune the pretrained T2I model with high-quality face images, enabling the pretrained T2I model, SDXL, to develop a prior for facial details. 
Utilizing this face-oriented generative diffusion prior, we introduced AuthFace for the blind face restoration task, achieving authentic face restoration. 
Additionally, we introduced the time-aware latent facial feature loss to further improve the robustness of restoration in key facial features. 

\noindent \textbf{Limitation and Future Work:}
The data collection process is labor-intensive, requiring significant human resources for manual filtering; we plan to automate this by developing an aesthetic-oriented image quality assessment network.
Regenerating unique facial details (e.g., moles, scars) that are absent in degraded inputs remains a challenge. As future work, we will explore reference-based face restoration to address this limitation.

\clearpage

\noindent \textbf{Acknowledgment:}
This paper is supported by the National Natural Science Foundation of China (NSF) under Grant
No. 62206069 (affiliated with Guangzhou HKUST Fok Ying Tung Research Institute) and the MOE AcRF Tier 1 SSHR-TG Incubator Grant FY24 under Grant No. RSTG7/24.

\bibliographystyle{ACM-Reference-Format}
\balance
\bibliography{sample-base}

\end{document}